\documentclass[sn-mathphys-ay]{sn-jnl}
\usepackage{amsmath,amssymb}
\usepackage{times}
\usepackage{url}
\usepackage{latexsym}
\usepackage{comment}
\usepackage[table]{xcolor}
\usepackage{booktabs}
\usepackage{geometry}
\usepackage{array}
\usepackage{enumitem}
\usepackage{tabularx}
\usepackage{graphicx}
\usepackage{amsmath}
\usepackage{pifont}
\usepackage[T1]{fontenc}
\usepackage[utf8]{inputenc}
\usepackage{natbib}

\usepackage[edges]{forest}

\title{Alternatives To Next Token Prediction In Text Generation - A Survey}

\author*[1]{\fnm{Charlie} \sur{Wyatt}}\email{charles.wyatt@student.unsw.edu.au}
\author[1]{\fnm{Aditya} \sur{Joshi}}\email{aditya.joshi@unsw.edu.au}
\author[1]{\fnm{Flora} \sur{Salim}}\email{flora.salim@unsw.edu.au}

\affil[1]{\orgdiv{School of Computer Science and Engineering},
          \orgname{UNSW Sydney},
          \orgaddress{\city{Sydney}, \country{Australia}}}
\date{}

\begin{document}
\maketitle

\begin{abstract}
The paradigm of Next Token Prediction (NTP) has driven the unprecedented success of Large Language Models (LLMs), but is also the source of their most persistent weaknesses such as poor long-term planning, error accumulation, and computational inefficiency. Acknowledging the growing interest in exploring alternatives to NTP, the survey describes the emerging ecosystem of alternatives to NTP. We categorise these approaches into five main families: (1) \textbf{Multi-Token Prediction}, which targets a block of future tokens instead of a single one; (2) \textbf{Plan-then-Generate}, where a global, high-level plan is created upfront to guide token-level decoding; (3) \textbf{Latent Reasoning}, which shifts the autoregressive process itself into a continuous latent space; (4) \textbf{Continuous Generation Approaches}, which replace sequential generation with iterative, parallel refinement through diffusion, flow matching, or energy-based methods; and (5) \textbf{Non-Transformer Architectures}, which sidestep NTP through their inherent model structure. By synthesizing insights across these methods, this survey offers a taxonomy to guide research into models that address the known limitations of token-level generation to develop new transformative models for natural language processing.
\end{abstract}

\section{Introduction}

Large Language Models (LLMs) have achieved remarkable performance across a wide range of natural language processing (NLP) tasks, from question answering and summarization to creative writing and coding~\cite{hendrycks2021measuringmassivemultitasklanguage, srivastava2023imitationgamequantifyingextrapolating, jiang2024surveylargelanguagemodels}. At the heart of this success lies a simple yet powerful objective: \textbf{Next Token Prediction (NTP)}~\cite{vaswani2023attentionneed}. Defined as the task of predicting the next subword token given the previous tokens in a sequence, NTP has become the dominant training and decoding paradigm for modern Transformer models.

Yet, despite their fluency, LLMs often exhibit brittle behavior: they hallucinate facts, lose coherence in long outputs \citep{maharana2024evaluatinglongtermconversationalmemory}, and struggle with tasks that require long-term planning or structured reasoning \citep{hao2023reasoninglanguagemodelplanning}. These limitations persist even as model scale, context length, and training data continue to grow. A growing body of work suggests that these failures are not incidental but symptomatic of deeper structural problems with the next-token prediction framework itself. As \citet{bachmann2024pitfallsnexttokenprediction} argue, the token-level chain rule that NTP relies on can break down, as the probability of the next most likely token does not guarantee the start of the most probable \emph{overall sequence}.

\begin{figure*}[t]
    \centering
    \includegraphics[width=1\textwidth]{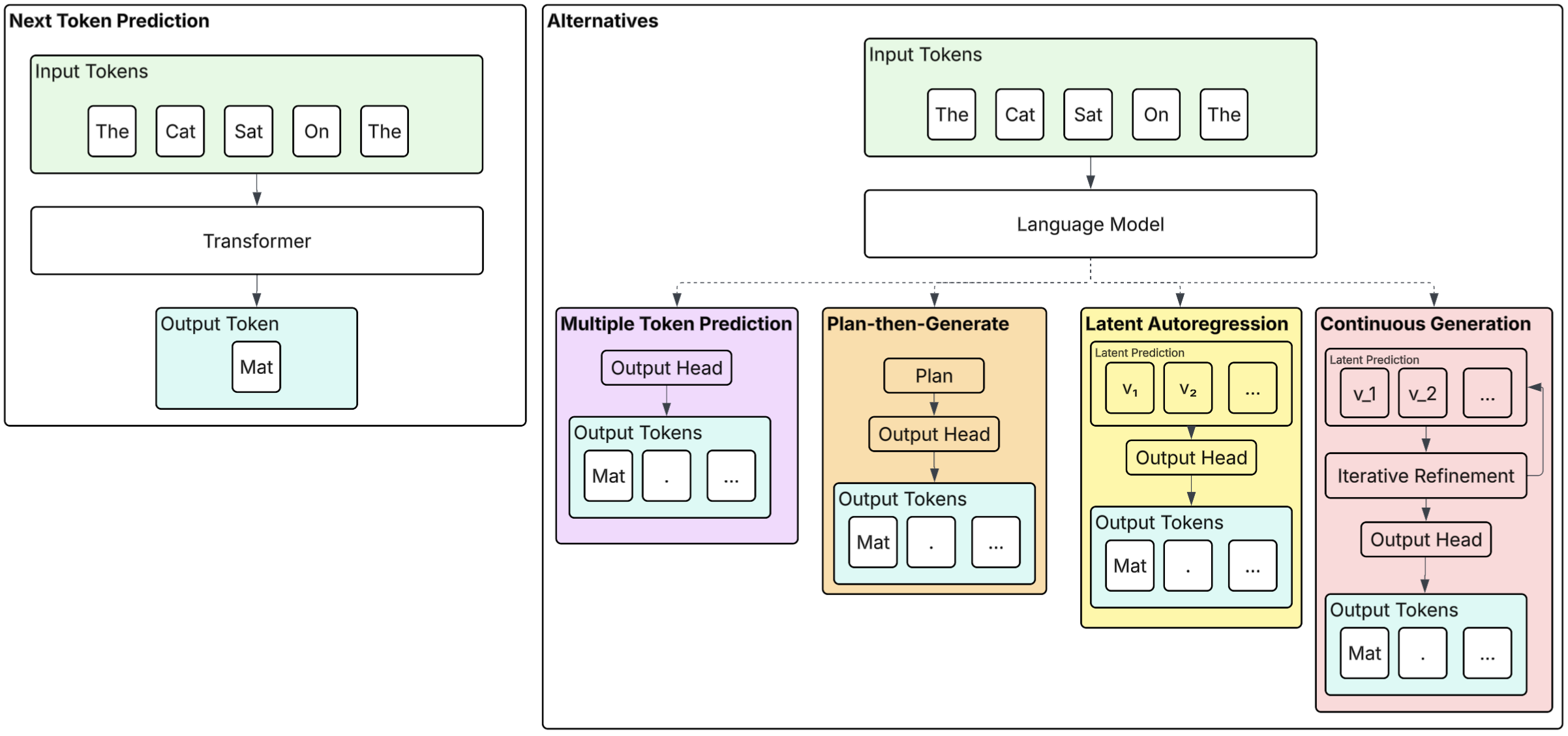}
    \caption{An abstracted visualization of Next Token Prediction and its alternatives}
    \label{fig:ntp_alternatives}
\end{figure*}

This survey \textit{examines the emerging ecosystem of alternatives to next-token prediction}. We argue that many core weaknesses of LLMs -  greedy decoding, error accumulation, misaligned supervision, granularity mismatches, and computational inefficiencies - are downstream of the NTP objective. As Figure~\ref{fig:ntp_alternatives} illustrates, a diverse set of methods now reimagine the prediction target, shifting away from token-by-token generation (as in the case of NTP) toward richer and more global semantic representations. While we organize these into five main families for clarity, emerging variants like flow matching and energy-based models represent important extensions within these categories, particularly in continuous generation approaches.
\section{Motivation}
This section outlines the core limitations of NTP and motivates the need for alternative decoding and training paradigms. As is evident from the discussion, the use of NTP poses challenges in several aspects of language modeling itself.

\subsection{Inference-Time Failures: Greedy Generation and Error Accumulation}

NTP is an \textbf{autoregressive} generation strategy, \textit{i.e.}, tokens are generated one at a time where the probability of the next token depends only on the existing context. Formally, given a sequence of tokens $x = (x_1, x_2, \dots, x_T)$, the model is trained to maximize the likelihood:

\begin{equation}
P(x) = \prod_{t=1}^{T} P(x_t \mid x_{<t}),
\end{equation}

where $x_{<t}$ denotes all tokens preceding the position $t$ and $x_t$ denotes the token at the position $t$. While this enables fluent local continuations, it also encourages \textbf{greedy generation}—a process that myopically optimizes for short-term token likelihood rather than long-range coherence. This greediness can lead to \textbf{error accumulation} \cite{rohatgi2025computationalstatisticaltradeoffsnexttokenprediction}, where minor inaccuracies early in generation compound into significant incoherence or factual drift later in the sequence.

These issues are especially pronounced in long-form generation tasks such as summarization and storytelling. As observed in surveys on abstractive summarization \citep{nnadi2024surveyabstractivetextsummarization}, even strong transformer models often struggle with maintaining global consistency and factual grounding across lengthy generations. Although retrieval-augmented generation and extended context windows mitigate some of these issues, they do not fundamentally address the underlying problem: generation proceeds \emph{locally}, without a representation of, or commitment to, a global plan.

Indeed, despite near-perfect performance on ``needle-in-a-haystack'' retrieval tasks (e.g., \textsc{Gemini 1.5}'s \(>99\%\) accuracy \citep{geminiteam2024gemini15unlockingmultimodal}), LLMs still underperform at generating coherent long-document summaries. As shown by \citet{maharana2024evaluatinglongtermconversationalmemory}, LLMs struggle with long-context understanding and temporal grounding, even when equipped with large context windows. A similar phenomenon has been documented outside pure text generation, where the model’s step-wise choices reveal a preference for short-term gains.
Recent studies in decision-making have further characterized this failure mode as \textbf{greediness in action selection}, where LLMs prematurely commit to locally optimal continuations rather than exploring alternate trajectories \citep{schmied2025llmsgreedyagentseffects}. This behaviour reflects a broader limitation in NTP-based generation: the inability to revise earlier decisions in light of later ones, or to reason over long-term structure during decoding.

\subsection{Granularity Mismatch: Tokens vs. Ideas}
Another potential shortcoming of NTP is its mismatch in granularity: \textbf{NTP operates over subword tokens, but human reasoning unfolds over ideas, sentences, and discourse structures}. Modern LLMs typically rely on subword tokenization, most commonly the Byte-Pair Encoding (BPE) algorithm \citep{sennrich2016neuralmachinetranslationrare}. While effective for handling large vocabularies, BPE fragments text into units that are neither linguistically natural nor semantically meaningful. In contrast, human communication is organized around higher-level structures such as phrases, sentences, and discourse segments. As a result, tasks like summarization or dialogue understanding—which require reasoning over ideas and maintaining global coherence—are forced to operate on token sequences that fail to reflect these natural semantic boundaries. 

This granularity misalignment also manifests in \textit{unexpected and undesirable behaviors} at the token level, such as the well-documented handling of `glitch tokens" like \texttt{solidgoldmagikarp} \citep{land2024fishingmagikarpautomaticallydetecting}, which arise from under-training or limited exposure in the corpus. Although recent work has proposed methods to detect and mitigate these under-trained tokens, the fact that such ad hoc fixes are needed suggests that sub-word tokens lack the right level of abstraction for language understanding. Beyond these outlier cases, it also leads to surprisingly poor performance on trivial character-level tasks. \citet{cosma2025strawberryproblememergencecharacterlevel} and \citet{fu2024largelanguagemodelsllms} demonstrate that tokenized language models can stumble on rudimentary character-counting tasks such as answering the question \textit{``How many r's are in strawberry?"}. Similarly, autoregressive models fail to plan ahead on even the simplest task: ask an LLM ``\textit{How many words are in your response?}" and it will almost always miscount. Since the model generates each word step-by-step, the model cannot see ahead and adjust its plan accordingly. %Non-autoregressive models, such as text diffusion, overcome this limitation by treating the entire output as a global object rather than a sequential stream.

These failures collectively indicate a structural mismatch: \textbf{subword tokens are too large for character-level reasoning, yet are also too fine-grained for sentence or discourse level abstraction}. This problem is particularly acute for typologically rich languages such as Indic languages, where subword tokenizers demonstrate significantly degraded performance \citep{thakur2025artbreakingwordsrethinking}. This reinforces the idea that tokens are neither universally the right size nor flexibly compositional for every type of reasoning.

\subsection{Compute Inefficiency: Quadratic Attention and the Cost of Token-Level Generation}

Beyond the representation concerns of subword-by-subword generation, NTP is also computationally inefficient. Transformer-based LLMs compute attention over the full input sequence at each layer and timestep, incurring a time and memory complexity of $\mathcal{O}(n^2)$ for sequences of length $n$ \citep{vaswani2023attentionneed}. When language is modeled at the subword token level, even modest-length documents can require hundreds or thousands of decoding steps—each step involving full forward passes over all previous tokens. This results in expensive and often redundant computations, especially during generation. The environmental impact of these inefficiencies in modern NLP architectures has also been reported \cite{ansar2024surveytransformersnlpfocus}.

Recent work has explored architectural innovations like sparse attention \citep{roy-etal-2021-efficient, Nguyen_2025}, linear transformers \citep{beltagy2020longformerlongdocumenttransformer}, and retrieval-based compression to reduce this overhead. A more principled solution lies in rethinking the granularity of representation itself. If models operated over higher-order units—such as phrases, sentences, or discourse segments—rather than atomic tokens, they could exploit the natural structure of language to reduce the input length into transformers and thereby reduce computational overhead. 

Recent architectural innovations have attempted to address these inefficiencies through modified attention mechanisms. Grouped Query Attention (GQA)~\citep{ainslie2023gqatraininggeneralizedmultiquery} and Multi-Query Attention (MQA)~\citep{shazeer2019fasttransformerdecodingwritehead} reduce memory requirements by sharing key-value heads across multiple query heads, though they still operate fundamentally at the token level. While these optimizations improve efficiency, they don't address the core granularity mismatch—models still process hundreds of tokens even for conceptually simple operations. The tight coupling between attention mechanisms and token-level processing suggests that moving beyond NTP may require rethinking attention itself.

Thus, alternatives to NTP are not only motivated by a desire for better reasoning and coherence, but also by the need for more efficient and scalable inference.

\subsection{Training-Time Failures: Misaligned Supervision in Teacher-Forcing}

A deeper issue arises during training. Models are typically trained using teacher-forcing, where they learn to predict the next token given the true previous tokens. As \citet{bachmann2024pitfallsnexttokenprediction} argue, this setup can cause models to learn \textbf{spurious shortcuts}, a phenomenon they call the “Clever Hans” effect. Instead of learning to plan, models exploit the revealed output prefix to guess plausible continuations without understanding the task. In lookahead tasks such as path-finding, this leads to \textbf{in-distribution failures}, where models trained to 100\% accuracy fail completely at test time. Even for trivial tasks, teacher-forced models struggle to learn the correct planning mechanism when supervision is misaligned.

\subsection{``Larger-than-token" representations exist inside Transformers}

After outlining the numerous failures induced by NTP, it might be tempting to conclude that the Transformer architecture itself is fundamentally flawed. However, a compelling body of evidence suggests the opposite: the problem may not be the core model, but the constraints imposed by its own token-by-token decoding. While a Transformer's \textit{output} is a simple, sequential stream of tokens, its \textit{internal} latent states exhibit a remarkable degree of sophistication and foresight.

Researchers have found that internal embeddings in LLMs naturally form ``larger-than-token" abstractions, with specific neurons or activation patterns corresponding to multi-token concepts, like people \citep{ghandeharioun2024patchscopesunifyingframeworkinspecting}, places \citep{templeton2024scaling}, or complex terms \citep{kaplan2025tokenswordsinnerlexicon}. This suggests that transformers learn to reason with high-level semantic units, even when only trained to predict the next token.

Furthermore, \textbf{Future Lens} \citep{Pal_2023}, has shown that a single hidden state can anticipate multiple future tokens with surprising accuracy. Using linear probes on models like GPT-J-6B, they show that middle-layer activations can predict two or three tokens ahead, proving that transformers implicitly build multi-step plans long before they are decoded. The model isn't as myopic as its output suggests; it's actively planning ahead.

In short, the limitations of NTP may not stem from a model's inability to represent complex ideas, but from the constraint of having to express them one subword at a time. The following sections survey methods that explicitly try to break this bottleneck.

\paragraph{Motivation Summary}
Together, these limitations point toward \textbf{a fundamental inadequacy in NTP as a general-purpose generative paradigm.} Consequently, this survey explores emerging alternatives—ranging from multi-token and latent-space decoding to non-autoregressive training—and outlines how they aim to bridge the gap between token-level generation and idea-level reasoning.

\begin{table*}[ht]
\small
\centering
\renewcommand{\arraystretch}{1.15}
\begin{tabularx}{\textwidth}{>{\raggedright\arraybackslash\bfseries}p{3cm} >{\raggedright\arraybackslash}p{4.5cm} >{\raggedright\arraybackslash}X}
\toprule
\rowcolor{gray!20}
\textbf{Category} & \textbf{Definition} & \textbf{Representative Papers} \\
\midrule
Multi-Token Prediction &
Generates several future tokens at once from some shared latent space ("trunk"). &
\begin{itemize}[noitemsep, left=0pt, topsep=0pt]
  \item \citet{qi2020prophetnetpredictingfuturengram} — ProphetNet
  \item \citet{gloeckle2024betterfasterlarge} — Better \& Faster LLMs via Multi-token Prediction
  \item \citet{ahn2025efficientjointpredictionmultiple} — Efficient Joint Prediction of Multiple Future Tokens
  \item \citet{deepseekai2025deepseekv3technicalreport} — DeepSeek-V3 Technical Report
\end{itemize} \\
\midrule
Plan-then-Generate &
Introduces a distinct, two-stage process where a global, high-level plan is generated upfront to guide subsequent token-by-token generation. &
\begin{itemize}[noitemsep, left=0pt, topsep=0pt]
  \item \citet{su2021planthengeneratecontrolleddatatotextgeneration} — Plan-then-Generate
  \item \citet{yin2024semformertransformerlanguagemodels} — Semformer
  \item \citet{lovelace2024diffusionguidedlanguagemodeling} — DGLM
  \item \citet{zhang2024plannergeneratingdiversifiedparagraph} — PLANNER
\end{itemize} \\
\midrule
Latent Reasoning &
Generates continuous latent vectors instead of, or alongside, discrete tokens. &
\begin{itemize}[noitemsep, left=0pt, topsep=0pt]
  \item \citet{hao2024traininglargelanguagemodels} — COCONUT
  \item \citet{geiping2025scalingtesttimecomputelatent} — Latent Reasoning
  \item \citet{lcmteam2024largeconceptmodelslanguage} — Large Concept Models
  \item \citet{tack2025llmpretrainingcontinuousconcepts} — CoCoMix
\end{itemize} \\
\midrule
Continuous Generation Approaches &
Denoises a noisy latent sequence over multiple steps instead of generating left-to-right. &
\begin{itemize}[noitemsep, left=0pt, topsep=0pt]
  \item \citet{li2022diffusionlmimprovescontrollabletext} — Diffusion-LM
  \item \citet{ye2025autoregressiondiscretediffusioncomplex} — Multi-Granularity Diffusion Modeling
  \item \citet{shabalin2025tencdmunderstandingpropertiesdiffusion} — TEncDM
  \item \citet{nie2025scalingmaskeddiffusionmodels} — Scaling Masked Diffusion
  \item \citet{arriola2025blockdiffusioninterpolatingautoregressive} — Block Diffusion
  \item Gemini Diffusion — \url{https://deepmind.google/models/gemini-diffusion/}
  \item LLaDA — \url{https://ml-gsai.github.io/LLaDA-demo/}
  \item \citet{lipman2023flowmatchinggenerativemodeling} — Flow Matching
  \item \citet{hu-etal-2024-flow} — Flow Matching for Text
  \item \citet{xu2025energybaseddiffusionlanguagemodels} — EDLM
  \item \citet{deng2020residualenergybasedmodelstext} — Residual EBMs
\end{itemize} \\
\midrule
Non-Transformer Architectures &
Models that inherently sidestep token-by-token autoregression through different core mechanics, such as recurrent states or joint embedding. &
\begin{itemize}[noitemsep, left=0pt, topsep=0pt]
  \item \citet{gu2024mambalineartimesequencemodeling} — Mamba
  \item \citet{assran2023selfsupervisedlearningimagesjointembedding} — I-JEPA
  \item Recurrent Neural Networks (RNNs/LSTMs)
  \item \citet{huang2025llmjepalargelanguagemodels} — LLM-JEPA

\end{itemize} \\
\bottomrule
\end{tabularx}
\caption{Taxonomy of alternatives to Next Token Prediction in transformers.}
\label{tab:ntp_alternatives}
\end{table*}

\section{Scope}

This survey was conducted through a systematic review of recent literature on alternatives to next-token prediction in language models. We searched major NLP venues including ACL, EMNLP, ICLR, NeurIPS, and ICML, as well as prominent preprint repositories (arXiv) covering the period from 2020 to 2025. We applied inclusion criteria focusing on methods that explicitly challenge or modify the standard next-token prediction paradigm—that is, approaches that move beyond predicting individual subword tokens to instead generate or predict "larger-than-token" units such as multiple tokens simultaneously, sentence-level representations, or global structural plans. We prioritized methods that fundamentally alter the generative process itself rather than optimizing within the existing token-by-token framework. We explicitly exclude attention mechanism variants such as Grouped Query Attention (GQA), Multi-Query Attention (MQA), and other architectural optimizations surveyed by \citet{Brauwers_2023} and \citet{roy-etal-2021-efficient}. While these innovations improve computational efficiency, they fundamentally operate within the NTP paradigm. Our focus is on methods that challenge the token-level generation paradigm itself, not those that optimize within it. Papers were excluded if they focused solely on computational efficiency improvements, retrieval augmentation or prompt engineering.

% Add a paragraph on how the papers were collected. Potential questions would be: did you use only ACL? did you select certain conferences only? How many papers did you have in the pool? Did you rule out any papers? What was the 'inclusion and exclusion criteria'?
Based on our survey, we categorise the alternatives to NTP into five broad families:

\begin{itemize}[leftmargin=1.5em]
    \item \textbf{Multi-Token Prediction (MTP)} changes the prediction target from a single token to a block of future tokens, typically using a shared context representation to feed multiple output heads.

    \item \textbf{Plan-then-Generate (PtG)} introduces a distinct, two-stage process where a global, high-level plan is generated upfront to guide the subsequent token-by-token generation.

    \item \textbf{Latent Reasoning (LR)} shifts the core autoregressive process into a continuous latent space, where the model's primary task is to generate an evolving sequence of latent states. These latent states are then used to predict the next sequence of tokens.

    \item \textbf{Continuous Generation Approaches (CG)} replace left-to-right autoregressive generation with parallel, iterative refinement processes over a \emph{global representation} of the entire output, including diffusion, flow matching, and energy-based methods. Instead of autoregressive token prediction, tokens are produced through multiple rounds of simultaneous refinement across all positions in the output window.

    \item \textbf{Non-Transformer Architectures (NTA)} employ alternative model backbones (e.g., State Space Models) whose inherent mechanics for sequence processing naturally sidestep the NTP paradigm by not requiring to generate tokens explicitly as a sequence.
\end{itemize}

\begin{figure*}[ht]
    \centering
    \includegraphics[width=0.5\textwidth]{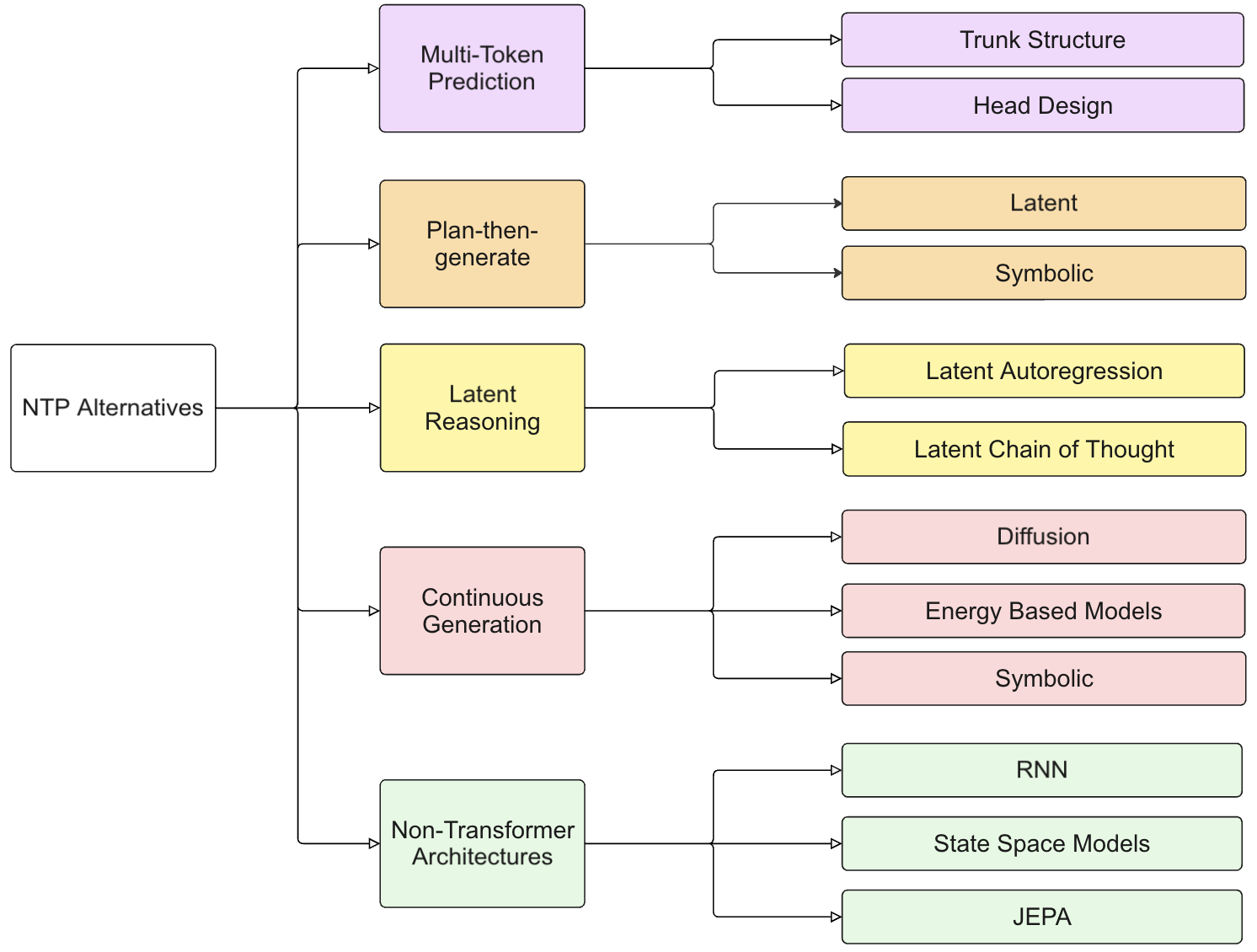}
    \caption{Taxonomy of alternatives to next-token prediction (NTP)}
    \label{fig:taxonomy}
\end{figure*}

Based on the families, we illustrate a taxonomy in Figure~\ref{fig:taxonomy}, and synthesizes insights from each method. Our goal is to provide a roadmap for researchers exploring how language models might move beyond token-level generation—toward models that can plan, reason, and communicate more like humans.

While survey papers exist for each of these families, to our knowledge, no paper has explicitly unified these into a common taxonomy within the scope of alternatives to NTP. Table~\ref{tab:scope-surveys} demonstrates that substantial prior work has examined individual categories—particularly Plan-then-Generate and Continuous Generation Approaches approaches—validating our taxonomic divisions while highlighting the gap in unified analysis across all five families. While these works validate the importance of individual approaches, they do not examine the common thread that unites them: the fundamental challenge to token-by-token generation that motivates our taxonomic framework.

\begin{table*}[ht]
\small
\centering
\renewcommand{\arraystretch}{1.2}
\begin{tabularx}{\textwidth}{>{\raggedright\arraybackslash\bfseries}p{9cm} >{\raggedright\arraybackslash}X}
\toprule
\rowcolor{gray!20}
\textbf{Paper Name} & \textbf{Category} \\
\midrule
Large Language Models for Planning: A Comprehensive and Systematic Survey \citep{cao2025largelanguagemodelsplanning} & Plan-then-Generate \\
\midrule
PlanGenLLMs: A Modern Survey of LLM Planning Capabilities \citep{wei2025plangenllmsmodernsurveyllm} & Plan-then-Generate \\
\midrule
A Survey of Diffusion Models in Natural Language Processing \citep{zou2023surveydiffusionmodelsnatural} & Continuous Generation Approaches \\
\midrule
Diffusion Models for Non-autoregressive Text Generation: A Survey \citep{li2023diffusionmodelsnonautoregressivetext} & Continuous Generation Approaches \\
\midrule
Energy-Based Models with Applications to Speech and Language Processing \citep{Ou_2024} & Continuous Generation Approaches \\
\midrule
A Survey on Latent Reasoning \citep{zhu2025surveylatentreasoning} & Latent Reasoning \\
\midrule
Next Token Prediction Towards Multimodal Intelligence: A Comprehensive Survey \citep{chen2024tokenpredictionmultimodalintelligence} & Multi-Token Prediction (sectional coverage) \\
\bottomrule
\end{tabularx}
\caption{Survey papers aligned with the families defined in this paper.}
\label{tab:scope-surveys}
\end{table*}

\section{Multi-token prediction}
\subsection{Definition}
\begin{figure}[h!]
    \centering
    \includegraphics[width=0.5\columnwidth]{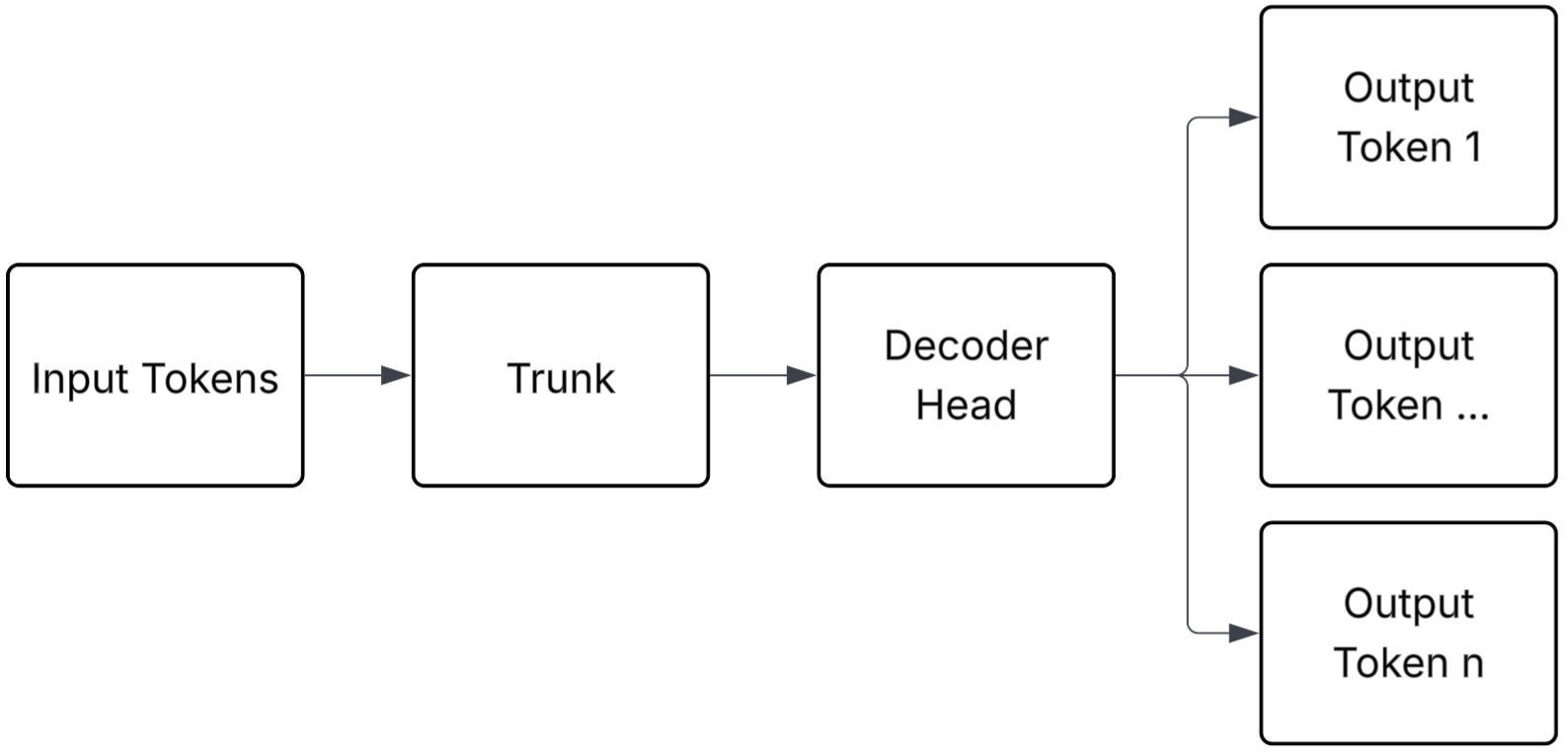}
    \caption{High-level architecture of a Multi-Token Prediction (MTP) model. A shared "trunk" processes the input context and feeds a representation to a decoder head, which generates multiple future tokens in parallel.}
    \label{fig:mtp-diagram}
\end{figure}

To accelerate inference and encourage short-term planning, \textbf{Multi-Token Prediction (MTP)} models change the prediction target from a single token to a block of \emph{k} future tokens. This is typically achieved by using a shared context representation (a ``trunk”) to feed multiple parallel output heads, each predicting a different future token. This aims to improve \textbf{computational efficiency} and \textbf{greedy generation}. Figure~\ref{fig:mtp-diagram} illustrates this architecture.

Formally, the probability is calculated as $$P(x_{t+1:t+k} \mid x_{\leq t}) = \prod_{i=1}^k P(x_{t+i} \mid h_t)$$ where the shared trunk representation is $h_t = f_{\text{trunk}}(x_{\leq t})$.

This setup breaks the strict chain rule of next-token prediction while also allowing for parallel decoding \citep{Santilli_2023}.

\subsection{Key Approaches}
Table~\ref{tab:mtp-comparison}) compares the key approaches for MTP in terms of the trunk, MTP head and number of predicted tokens.  We now present these approaches in detail.
\citet{qi2020prophetnetpredictingfuturengram} first popularized this strategy with ProphetNet, which uses an n-stream self-attention mechanism to predict the next \(n\) tokens in parallel from a shared context stream (trunk). Each stream acts as an MTP head conditioned on the same internal state, enabling long-range planning while maintaining compatibility with standard inference. 

\citet{gloeckle2024betterfasterlarge} extend this idea from encoder-decoder models like ProphetNet to large-scale autoregressive LLMs by training a standard Transformer (trunk) with multiple output heads, each predicting a separate future token. Unlike ProphetNet, which uses offset attention streams, these heads are trained in parallel from a shared representation and used only during training unless speculative decoding is enabled. Despite its simplicity, this method improves sample efficiency and enables speculative decoding without architectural changes at inference. Notably, MTP heads are trained jointly, but only the next-token head is used at test time. Their 13B model achieves up to 17\% higher accuracy on MBPP and a 3× speedup in decoding, while also fostering stronger induction capabilities and reasoning.

\citet{ahn2025efficientjointpredictionmultiple} propose Joint Token Prediction (JTP), which improves upon prior MTP variants by modeling the joint distribution of future tokens rather than treating them as conditionally independent. Like previous methods, JTP uses a shared Transformer trunk, but introduces a lightweight attention module (``Fetch”) to process teacher-forced tokens and refine the trunk’s output before feeding it to the MTP head. This bottleneck forces the trunk to encode richer multi-step planning signals. JTP outperforms other MTP methods on synthetic reasoning benchmarks with fewer parameters and minimal computational overhead.

MTP is commonly used as an auxiliary objective during training. \citet{deepseekai2025deepseekv3technicalreport} show that adding MTP heads during training—and discarding them at inference—improves generalization across benchmarks, even in 200B+ models. Here, the shared Transformer trunk receives gradients from multiple future-token heads, encouraging broader temporal representations. When retained at inference, these heads enable speculative decoding: DeepSeek reports that 85–90\% of the speculative proposed tokens are accepted, significantly accelerating generation with minimal degradation in quality.

\begin{table*}[ht]
\centering
\small
\renewcommand{\arraystretch}{1.2}
\begin{tabular}{>{\bfseries}p{3cm} p{4cm} p{4cm} p{3cm}}
\toprule
Model & Trunk & MTP Head(s) & Number of Predicted Tokens \\
\midrule
ProphetNet & Transformer encoder (seq2seq; encoder is shared trunk) & Offset decoder streams for future token prediction & 1 - 3 \\
Gloeckle et al. & Transformer trunk & Independent heads for each future token & 1 - 32 \\
JTP (Ahn et al.) & Transformer trunk & Fetch module + joint head for joint token prediction & 1 - 8 \\
DeepSeek-v3 & Transformer trunk & MTP heads (discarded or used for speculative decoding) & 1 - 2 \\
\bottomrule
\end{tabular}
\caption{Comparison of key MTP approaches in terms of the trunk structure, head design, and distinguishing features.}
\label{tab:mtp-comparison}
\end{table*}

\subsection{Failure Modes}
MTP commonly fails at the following:
\begin{itemize}[leftmargin=1.5em, noitemsep]
    \item \textbf{Limited planning depth:} Despite modeling multiple tokens, MTP often focuses on short-term future steps, lacking explicit mechanisms for deeper hierarchical or global planning.
    \item \textbf{Architectural inflexibility:} Integrating joint prediction heads (e.g., JTP’s Fetch module) may complicate training and require additional tuning to balance losses across heads.
    \item \textbf{Speculative decoding tradeoff:} When used at inference, speculative decoding introduces a trade-off between speed, accuracy and cost.
\end{itemize}

\subsection{Discussion}

Multi-token prediction (MTP) offers a simple, compatible extension to next-token prediction by forecasting blocks of future tokens instead of one at a time. This enables short-term planning and partial parallelization, addressing NTP’s sequential bottleneck without major architectural changes.

However, these models often only look 4-5 tokens into the future, and require new training setups that are not easily transferred to current models.

Overall, MTP improves efficiency and token-level foresight but lacks mechanisms for truly global planning or long-range structure.

\section{Plan-then-Generate}
\label{sec:plan-then-generate}

\subsection{Definition}
\begin{figure}[h!]
    \centering
    \includegraphics[width=0.5\columnwidth]{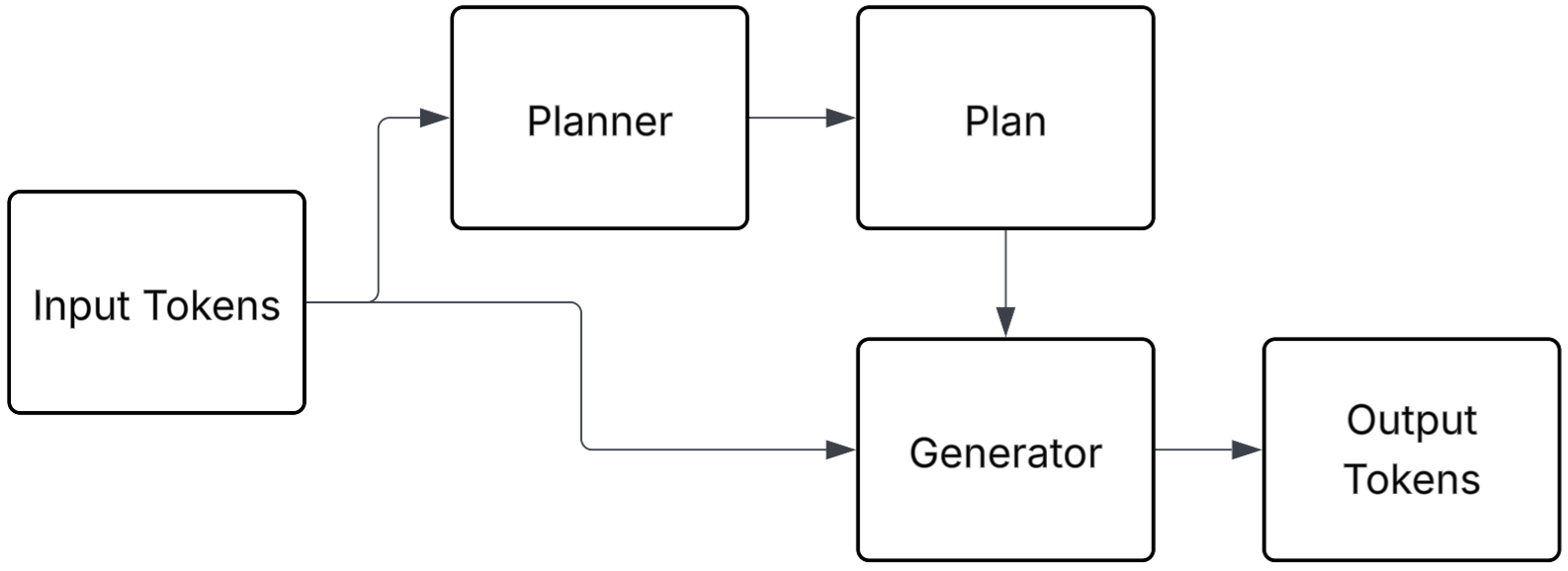}
    \caption{The PtG framework. A high-level plan is created, which is then used alongside the original input to guide the generator in producing the final tokens.}
    \label{fig:ptg-diagram}
\end{figure}

While MTP retains NTP's original formulation of generating a sequence one token at a time, plan-then-generate offers an alternative. Instead of generating text greedily from left to right, \textbf{Plan-then-Generate (PtG)} models introduce a high-level planning stage to guide subsequent token generation. This two-stage process is depicted in Figure ~\ref{fig:ptg-diagram}. Broadly, PtG systems can be categorized as \textbf{symbolic}—which generate structured, interpretable plans—or \textbf{latent}, which learn abstract, continuous representations of global intent.

The generation process has two stages:

\begin{itemize}[leftmargin=1em]
    \item \textbf{Planning:} Generate a high-level abstract plan $z$ from input context $c$
    \item \textbf{Generation:} Generate the output sequence $y$ conditioned on $c$ and $z$
\end{itemize}

Or, formally -
\begin{equation}
z \sim P_{\text{planner}}(z \mid c)
\label{eq:planning-stage}
\end{equation}
\begin{equation}
y \sim P_{\text{generator}}(y \mid z, c) = \prod_{i=1}^{N} P(y_i \mid y_{<i}, z, c)
\label{eq:generation-stage}
\end{equation}

We categorize PtG systems along three key axes:

\begin{itemize}[leftmargin=1em]
  \item \textbf{Hierarchy Unit:} What level of abstraction the plan operates on (e.g., slot, sentence, paragraph)
  \item \textbf{Planner Module:} How the plan is created (e.g., CRF, diffusion, variational autoencoder)
  \item \textbf{Conditioning Strategy:} How the plan influences decoding
\end{itemize}

\paragraph{Conditioning Strategies.} The \textit{conditioning strategy} defines how the generated plan is used to steer generation:
\begin{itemize}[leftmargin=1.5em]
    \item \textbf{Prepending latent tokens:} Learned embeddings are inserted as special tokens between prompt and output (e.g., Semformer).
    \item \textbf{Cross-attention to plan:} The decoder attends directly to the symbolic plan, allowing fine-grained alignment (e.g., PlanGen).
    \item \textbf{Soft prompt from embedding:} The plan is mapped into soft prompt vectors that bias decoding (e.g., DGLM).
    \item \textbf{Adapter-based injection:} Plan vectors are fed into lightweight adapter modules within a frozen decoder (e.g., Cornille et al.).
    \item \textbf{Per-layer latent fusion:} At each transformer layer, a fixed plan vector is projected and concatenated to every token’s hidden state, providing persistent global conditioning throughout decoding (e.g., PLANNER).
\end{itemize}

\vspace{1em}
\subsection{Representative Papers}
\subsubsection*{Symbolic Planning Models}

Early PtG work in structured data-to-text generation~\citep{puduppully2021datatotextgenerationmacroplanning, chen2020neuraldatatotextgenerationdynamic, li2023planthenseamefficienttabletotextgeneration} employed symbolic plans to select and order information before realization. \textbf{PlanGen}~\citep{su2021planthengeneratecontrolleddatatotextgeneration} predicts a sequence of content slots from RDF triples using a CRF, which a BART model then verbalizes. \textbf{LLM+P}~\citep{liu2023llmpempoweringlargelanguage} generalizes this idea to classical planning: an LLM formalizes a task into a structured PDDL program, which an external planner solves before the LLM verbalizes the result.

These systems exemplify \textit{neuro-symbolic} PtG: interpretable, modular, and structurally constrained, but limited by the need for symbolic input formats and task-specific schemas.

\vspace{1em}
\subsubsection*{Latent Planning Models}

Recent PtG models replace symbolic plans with learned \textbf{latent representations}, \textit{i.e.}, continuous vectors or discrete embeddings that encode global intentions without requiring hand-crafted structure.

\paragraph{Semformer}~\citep{yin2024semformertransformerlanguagemodels} learns special tokens that reconstruct sentence embeddings, inserted between prompt and output to provide lightweight planning signals.

\paragraph{DGLM}~\citep{lovelace2024diffusionguidedlanguagemodeling} samples an abstract intent vector from a diffusion model over Sentence-T5 latents, which is transformed into a soft prompt that conditions decoding.

\paragraph{PLANNER}~\citep{zhang2024plannergeneratingdiversifiedparagraph} encodes an entire paragraph into a latent vector using a VAE and denoises it via diffusion. The final plan conditions a decoder to improve structure and diversity.

\paragraph{Cornille et al.}~\citep{cornille2024learningplanlanguagemodeling} cluster sentence embeddings into discrete writing actions. These serve as modular planning units and are injected into a frozen LLM via adapter layers. Externally trained planners outperform internal ones in structuring long-form outputs.

These methods retain the abstraction benefits of symbolic planning while improving scalability and flexibility.

\paragraph{Broader Applications.} While this survey focuses on PtG for text generation, similar strategies have been applied to multi-agent planning~\citep{wang2024cooperativestrategicplanningenhances} and retrieval-augmented generation~\citep{lee2024planragplanthenretrievalaugmentedgeneration,lyu2024retrieveplangenerationiterativeplanninganswering}. These extensions are outside the scope of this review but demonstrate PtG's growing influence beyond generation.

\begin{table*}[ht]
\centering
\small
\renewcommand{\arraystretch}{1.2}
\begin{tabular}{>{\bfseries}p{3.3cm} p{2.8cm} p{4cm} p{4cm}}
\toprule
Model & Hierarchy Unit & Planner Module & Conditioning Strategy \\
\midrule
PlanGen~\citep{su2021planthengeneratecontrolleddatatotextgeneration} & Slot-level (symbolic) & CRF over RDF triples & Cross-attention to plan \\
Semformer~\citep{yin2024semformertransformerlanguagemodels} & Sentence-level (latent) & Autoencoder (offline) & Prepending latent tokens \\
DGLM~\citep{lovelace2024diffusionguidedlanguagemodeling} & Sentence-level (latent) & Diffusion over Sentence-T5 & Soft prompt from embedding \\
PLANNER~\citep{zhang2024plannergeneratingdiversifiedparagraph} & Paragraph-level (latent) & VAE + latent diffusion & Decode conditioned on latent \\
Cornille et al.~\citep{cornille2024learningplanlanguagemodeling} & Sentence-level (discrete actions) & Clustering + external classifier & Adapter-based plan injection \\
\bottomrule
\end{tabular}
\caption{Comparison of Plan-then-Generate models by hierarchy unit, planning method, and conditioning strategy.}
\label{tab:plan-then-generate-taxonomy}
\end{table*}

These methods scale PtG to open-domain settings but often sacrifice interpretability and simplicity in exchange.

\subsection{Failure Modes}
Despite architectural elegance, the following vulnerabilities have been reported for current PtG methods:
\begin{itemize}[leftmargin=1em]
    \item \textbf{Planner-Generator Misalignment:} If the plan is poor or inconsistent with the generation objective, downstream outputs degrade significantly.
    \item \textbf{Opaque Planning Representations:} Latent planners are not easily interpretable or debuggable, making it difficult to identify errors.
    \item \textbf{Rigidity of Symbolic Plans:} Symbolic methods require structured inputs and fail to generalize across domains.
    \item \textbf{Pipeline Complexity:} Most PtG systems require multiple pretrained components (e.g., VAE, diffusion models, clustering pipelines), increasing training cost and brittleness.
\end{itemize}

\subsection{Discussion}
Plan-then-Generate offers a promising strategy to introduce global coherence and task structure into generation, mitigating the myopia of next-token decoding. However, current implementations introduce steep engineering and alignment costs. Symbolic planners lack scalability, while latent planners introduce interpretability and reliability challenges. To become practical, PtG must evolve toward lighter, robust planning modules with end-to-end trainability and alignment guarantees.

\section{Latent Reasoning}

\subsection{Definition}
\textbf{Latent Reasoning (LR)} models replace token-level generation with autoregression over latent states, shifting the core generative loop into a continuous space. This addresses the \textbf{granularity mismatch} between subword tokens and high-level semantics, allowing models to operate at the level of sentences, thoughts, or concepts.

We group LR methods into two categories.

\subsubsection*{Latent Autoregression}
\label{sec:latent-autoregression}
These models generate a sequence of latent vectors and decode them into text afterward:

\begin{equation}
P(z \mid x) = \prod_{i=1}^{N'} P(z_i \mid z_{<i}, f_{\text{enc}}(x)) 
\end{equation}

\begin{equation}
P(y \mid z) = \prod_{i=1}^{N'} P(y_i \mid z_i)
\end{equation}

\textbf{SentenceVAE}~\citep{an2024sentencevaeenablenextsentenceprediction} encodes each sentence into a latent vector and autoregressively predicts the next. A decoder reconstructs each latent into a sentence. Compared to token-level baselines, it improves inference speed (2–3×), perplexity (46–75\%), and memory use (86–91\% reduction).

\textbf{Large Concept Models (LCMs)}~\citep{lcmteam2024largeconceptmodelslanguage} scale this approach, generating one ``concept'' vector per sentence to enable long-range abstraction and reduce sequence length.

\textbf{CoCoMix}~\citep{tack2025llmpretrainingcontinuousconcepts} interleaves latent concept vectors (from a pretrained autoencoder) with token decoding, enriching each step with high-level semantic context. This improves sample efficiency and performance on reasoning tasks.

\subsubsection*{Latent Chain-Of-Thought}
Here, the model refines a latent state over several internal steps before emitting a token. The steps indicate the chain of thought.

\begin{equation}
h_i^{(r)} = f_{\text{core}}(h_i^{(r-1)}, x) 
\end{equation}

\begin{equation}
P(y_i \mid y_{<i}, x) = f_{\text{out}}(h_i^{(R)})
\end{equation}

\textbf{Latent Reasoning}~\citep{geiping2025scalingtesttimecomputelatent} introduces an inner loop that applies a fixed ``core" module multiple times before token generation. This allows more computation without increasing model size, scaling inference along a new axis: internal iteration.

\textbf{Coconut}~\citep{hao2024traininglargelanguagemodels} performs latent reasoning between text segments, enabling breadth-first exploration of output trajectories before committing to a token sequence.

\vspace{0.5em}
Together, these models move beyond discrete decoding, enabling generation over semantic units like sentences, thoughts, or concepts.

\begin{table*}[ht]
\centering
\small
\renewcommand{\arraystretch}{1}
\begin{tabular}{>{\bfseries}p{2cm} >{\bfseries}p{2cm} p{2.5cm} p{4.5cm} p{2.5cm}}
\toprule
Model & Mechanism & Token Unit & Key Distinction & Output Format \\
\midrule
SentenceVAE & Latent Autoregression & Sentence embedding & Autoregresses over sentence vectors with a base LLM decoder & Discrete (sentences) \\
LCM & Latent Autoregression & Sentence embedding & Fully integrated latent autoregressive architecture & Discrete (sentences) \\
\midrule
CoCoMix & Latent Autoregression & Latent concept vector & Predicts token and concept vector at every step & Discrete (tokens) \\
\midrule
Latent Reasoning & Latent Chain of Thought & Continuous thought & Internal loop increases compute before each token & Discrete (tokens) \\
Coconut & Latent Chain of Thought & Continuous thought & Latent planning between text chunks & Mixed \\
\bottomrule
\end{tabular}
\caption{Comparison of Latent Reasoning Models.}
\label{tab:latent-dynamics-models}
\end{table*}

\subsection{Failure Modes}

Despite enabling generation over higher-level abstractions, latent reasoning models face distinct limitations:

\begin{itemize}[leftmargin=1em]
    \item \textbf{Opaque Generation Process:} Autoregression occurs in latent space, making it difficult to interpret, debug, or steer generation.
    \item \textbf{Latent Misalignment:} Poorly trained or misaligned latent encoders/decoders lead to degraded outputs or semantic drift during decoding.
    \item \textbf{Training Complexity:} Many models (e.g., SentenceVAE, LCM, CoCoMix) rely on complex multi-stage training involving external encoders, autoencoders, or diffusion priors.
    \item \textbf{Evaluation Instability:} Outputs often require post-decoding reconstruction, introducing noise that complicates evaluation and benchmarking.
    \item \textbf{Limited Robustness:} Out-of-distribution latent vectors (e.g., hallucinated concepts) may decode to nonsensical or unsafe outputs without explicit token-level grounding.
\end{itemize}

\subsection{Discussion}
Latent Reasoning offers a compelling departure from next-token prediction by shifting generation to a space of abstract semantic units. These methods achieve greater speed and generalization by reducing sequence length and operating at coarser granularity. However, they introduce new difficulties around interpretability, training complexity, and latent stability. Without token-level grounding, these systems can be fragile to errors in latent encoding, and their benefits are often offset by costly multi-stage pipelines. Future work should explore tighter integration between latent representations and discrete outputs, improving transparency and robustness while retaining semantic abstraction.

\section{Continuous Generation Approaches}
\label{sec:continuous-generation}
\subsection{Definition}
Continuous Generation approaches fundamentally depart from sequential token production by treating text generation as a global optimization problem. Rather than producing tokens one-by-one (whether in token space or latent space), these methods initialize the entire output simultaneously and refine it through iterative transformations. This paradigm shift from \textbf{sequential} to \textbf{parallel} generation enables three key advantages: (1) global planning and revision, (2) mitigation of error accumulation, and (3) bidirectional refinement where later context can influence earlier positions.

All continuous generation methods share a common framework: they define a trajectory from noise (or random initialization) to coherent text through multiple refinement steps. The specific mechanism varies—diffusion uses stochastic denoising, flow matching learns deterministic paths, and energy-based models follow gradient descent in an energy landscape—but the core principle remains: \textbf{generate everything at once, then refine globally}.

Note that while Latent Autoregression \ref{sec:latent-autoregression} shifts generation to a continuous space, it maintains the sequential, left-to-right generation paradigm. This distinguishes it from the approaches in this section, which abandon sequential processing entirely in favor of parallel refinement.

\subsection{Diffusion-Based Models}
Diffusion represents the most established continuous generation approach, treating generation as an iterative denoising process. Instead of emitting one token at a time, these models start with a noisy version of the full output and refine it over multiple steps.

\begin{figure*}[ht]
    \centering
    \includegraphics[width=\textwidth]{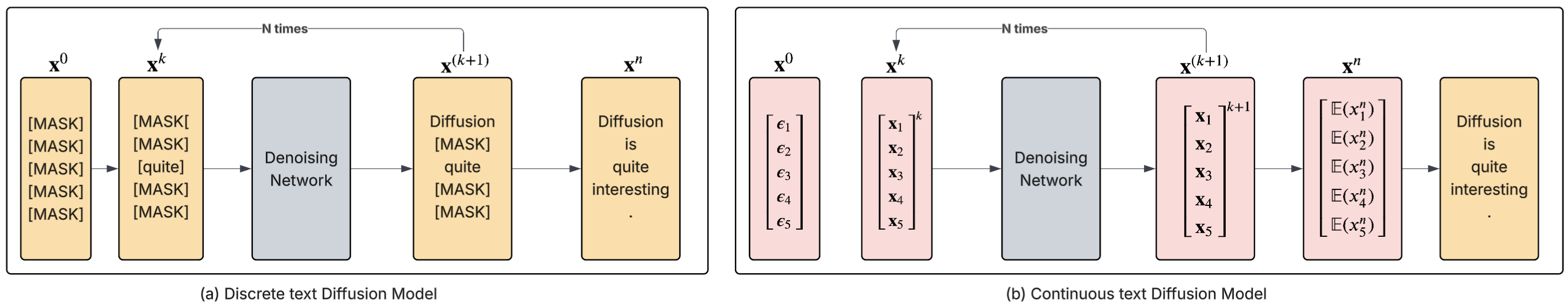}
    \caption{Comparison of text diffusion models. (a) A \textbf{Discrete Text Diffusion Model} starts with masked tokens and iteratively fills them in over N steps. (b) A \textbf{Continuous Text Diffusion Model} starts with random noise in an embedding space and iteratively denoises the latent vectors, which are then decoded into the final text.}
    \label{fig:diffusion-diagram}
\end{figure*}

However, adapting diffusion to language is nontrivial due to the discrete nature of text. Figure ~\ref{fig:diffusion-diagram} provides a conceptual overview of the two main approaches:
\begin{itemize}[leftmargin=1em]
  \item \textbf{Continuous diffusion} — Operates in embedding or latent spaces.
  \item \textbf{Discrete diffusion} — Operates over tokens or logits via masking and corruption.
\end{itemize}

A dedicated survey by \citet{li2023diffusionmodelsnonautoregressivetext} reviews the technical landscape of text diffusion models, covering noise schedules, conditioning, and integration with pretrained LMs. 

Table~\ref{tab:continuous-generation-models} categorizes recent Continuous Generation approaches along several dimensions.
\subsection{Representative Papers}

\subsubsection*{Continuous Diffusion Models}

These models apply noise to continuous representations and train a denoiser to recover the original signal.

\textbf{Diffusion-LM}~\citep{li2022diffusionlmimprovescontrollabletext} generates word embeddings via iterative denoising, allowing fine-grained control over syntax and semantics.

\textbf{TEncDM}~\citep{shabalin2025tencdmunderstandingpropertiesdiffusion} denoises in a frozen encoder’s latent space (e.g., BERT), capturing sentence-level semantics more effectively with fewer steps.

\subsubsection*{Discrete Diffusion Models}

Discrete models operate directly on tokens or logits, generalizing masked language modeling into a multi-step refinement process.

\textbf{MGDM}~\citep{ye2025autoregressiondiscretediffusioncomplex} improves symbolic reasoning by denoising from multiple noisy views, mitigating subgoal imbalance in AR models.

\textbf{Block Diffusion}~\citep{arriola2025blockdiffusioninterpolatingautoregressive} blends diffusion with autoregression by refining fixed-length blocks. This supports scalable, partially parallel decoding with token caching.

\textbf{Mercury}~\citep{labs2025mercuryultrafastlanguagemodels} scales discrete diffusion to commercial deployment, achieving \textbf{1100 tokens/sec} with latencies under \textbf{25ms}. Its coarse-to-fine, parallel denoising rivals top AR models in quality, offering a viable alternative for real-time coding and FIM tasks.

\subsubsection*{Emerging Directions}

Large-scale efforts like \textbf{Gemini Diffusion}\footnote{\url{https://deepmind.google/models/gemini-diffusion/}} and \textbf{LLaDA}\footnote{\url{https://ml-gsai.github.io/LLaDA-demo/}} signal growing industry investment in diffusion-based language modeling, especially for multimodal and instruction-tuned use cases.

\subsection{Flow Matching Models}

Flow matching~\citep{lipman2023flowmatchinggenerativemodeling} offers an alternative to diffusion that learns deterministic trajectories between noise and data distributions. Unlike diffusion's stochastic paths, flow matching constructs optimal transport maps that can be more sample-efficient. The framework has been successfully extended beyond its original formulation to simulation-based inference~\citep{dax2023flowmatchingscalablesimulationbased} and structured data like graphs~\citep{eijkelboom2025variationalflowmatchinggraph}. Recent work has specifically adapted flow matching to text generation~\citep{hu-etal-2024-flow}, treating sentences as points in a continuous semantic space and learning smooth transformations between random initialization and meaningful text. This approach entirely sidesteps tokenization by operating in embedding space throughout the generation process.

\subsection{Energy-Based Models}

Energy-based models (EBMs) define generation through an energy landscape where valid outputs correspond to low-energy configurations, fundamentally departing from token-by-token generation. Originally developed in statistical physics and later adopted in machine learning for vision tasks, EBMs learn a scalar energy function that assigns compatibility scores to entire configurations. Recent adaptations to NLP demonstrate multiple pathways for incorporating energy functions into language generation. \textbf{Energy-based Diffusion Language Models (EDLM)}~\citep{xu2025energybaseddiffusionlanguagemodels} introduce EBM formulations within discrete diffusion models, enabling sequence-level energy modeling at each diffusion step while leveraging pretrained LLMs through noise contrastive estimation. \textbf{Energy-Based Reward Models (EBRM)}~\citep{lochab2025energybasedrewardmodelsrobust} enhance LLM alignment by replacing scalar rewards with energy functions that better capture distributional uncertainty in human preferences, integrating seamlessly with existing alignment pipelines. \textbf{Energy-Based Transformers (EBT)}~\citep{gladstone2025energybasedtransformersscalablelearners} reframe the transformer architecture itself as an EBM, learning energy functions to assess compatibility between context and generated text. Finally, autoregressive variants like \textbf{Residual EBMs}~\citep{deng2020residualenergybasedmodelstext} impose energy-based post-processing on pretrained LLM outputs through contrastive objectives. These approaches collectively demonstrate that holistic sequence scoring via energy functions offers a principled alternative to the local, incremental decisions of NTP.

\begin{table*}[ht]
\centering
\small
\renewcommand{\arraystretch}{1.2}
\begin{tabular}{>{\bfseries}p{3cm} p{2.5cm} p{3cm} p{2.5cm} p{2.5cm}}
\toprule
Model & Approach Type & Latent Space & Granularity & Update Mechanism \\
\midrule
Diffusion-LM & Diffusion & Word embeddings & Token-level & Stochastic denoising \\
TEncDM & Diffusion & Frozen encoder latents & Sentence-level & Stochastic denoising \\
MGDM & Diffusion & Token logits & Token-level & Discrete denoising \\
Block Diffusion & Diffusion & Token blocks & Block-level & Block-wise refinement \\
Mercury & Diffusion & Token logits & Token-level & Parallel denoising \\
\midrule
FlowSeq & Flow Matching & Text embeddings & Sequence-level & Deterministic flow \\
\midrule
EDLM & Energy-Based & Token logits & Token-level & Energy minimization \\
EBRM & Energy-Based & Reward space & Sequence-level & Gradient descent \\
Residual EBMs & Energy-Based & Output logits & Token-level & Contrastive energy \\
\bottomrule
\end{tabular}
\caption{Comparison of Continuous Generation Approaches by type, latent space, granularity, and update mechanism.}
\label{tab:continuous-generation-models}
\end{table*}

\subsection{Failure Modes}
\begin{itemize}[leftmargin=1em]
    \item \textbf{Quality gap vs. AR models:} Despite improvements, most continuous generation models underperform strong autoregressive baselines in terms of fluency and coherence. Diffusion models require heavy tuning or distillation; flow matching models like FlowSeq show competitive but not superior performance; and energy-based models often struggle with mode coverage and sampling quality.
    
    \item \textbf{Inference cost and parallelism tradeoff:} While continuous generation is inherently parallelizable, the iterative refinement process incurs significant costs. Diffusion models require dozens of denoising steps; flow matching needs multiple ODE solver steps despite being deterministic; and energy-based models require expensive MCMC sampling or gradient-based optimization. Without architectural optimizations, all are slower than autoregressive decoding.
    
    \item \textbf{Fixed-length limitations:} Most continuous generation models struggle with variable-length outputs. Diffusion models like Diffusion-LM and MGDM require padding or truncation; flow matching operates on fixed-dimensional embeddings; and energy-based models typically define energy over fixed-size configurations.
    
    \item \textbf{Poor interpretability:} Generation trajectories are opaque across all approaches—diffusion follows stochastic paths in embedding space, flow matching traces deterministic but uninterpretable curves, and energy-based models descend abstract energy landscapes with no clear reasoning traces.
    
    \item \textbf{Complex training pipelines:} These models require specialized components that increase implementation overhead. Diffusion needs noise schedulers and denoising objectives; flow matching requires ODE solvers and transport cost optimization; energy-based models need contrastive sampling strategies and careful initialization to avoid mode collapse.
\end{itemize}

\subsection{Discussion}
Continuous Generation approaches break from the rigidity of token-by-token autoregressive generation by enabling global, parallel reasoning and more flexible generation paths. Each approach offers distinct advantages: Diffusion-LM supports rich control signals, while discrete methods like Mercury scale to production-level throughput; flow matching offers deterministic, theoretically grounded trajectories that can be more sample-efficient than stochastic diffusion; and energy-based models provide a principled framework for incorporating global constraints and preferences through holistic sequence scoring. Yet the cost of these benefits is high: slow inference, interpretability challenges, and complex training procedures remain core obstacles.

\section{Non-Transformer Architectures}

\subsection{Definition}
While previous approaches modify the generative objective within Transformers, this category targets a deeper shift: changing the model architecture itself. Non-Transformer Architectures (NTA) bypass the limitations of NTP by adopting fundamentally different sequence modeling mechanisms.

\subsection{Representative Papers}

\subsubsection*{Recurrent Neural Networks}

Recurrent Neural Networks (RNNs) were the state-of-the-art NLP architecture before the Transformer \citep{mikolov10_interspeech}. These models — including LSTMs \citep{sundermeyer12_interspeech} and GRUs \citep{chung2014empiricalevaluationgatedrecurrent} — process input sequentially, maintaining a hidden state that summarizes prior context. Unlike transformers, which compute attention over the entire sequence, RNNs make predictions using only this evolving summary. This results in linear time complexity and an implicit latent-space abstraction over all observed inputs.

However, RNNs struggle with vanishing gradients and limited memory, making them hard to scale for long-range dependencies. Recent work revisits this paradigm with more scalable variants, such as state space models.

\subsubsection*{State Space Models}

State Space Models (SSMs) improve on RNNs by tracking sequence information using continuous signals instead of discrete steps. Modern implementations like \textbf{Mamba}~\citep{gu2024mambalineartimesequencemodeling} achieve linear-time inference while retaining strong modeling capacity.

SSMs address key Transformer NTP bottlenecks:

\begin{itemize}[leftmargin=1.5em, noitemsep]
    \item \textbf{Efficiency:} Linear scaling with sequence length avoids $\mathcal{O}(n^2)$ attention cost.
    \item \textbf{Abstraction:} Continuous state evolution allows modeling over higher-level semantic units.
\end{itemize}
However, they remain underexplored in open-ended generation and bring their own challenges:

\begin{itemize}[leftmargin=1.5em, noitemsep]
    \item \textbf{Limited adoption:} Few large-scale applications in general-purpose NLP.
    \item \textbf{Opaque internals:} Dynamics are difficult to inspect or interpret.
    \item \textbf{Architectural fragility:} Sensitive to hyperparameters and discretization schemes.
\end{itemize}

While SSMs rethink temporal modeling, the next category, JEPAs, rethinks \textbf{prediction itself}—moving away from token-level outputs entirely.

\subsubsection*{Joint Embedding Predictive Architectures}

A more radical departure from NTP is offered by \textbf{Joint Embedding Predictive Architectures (JEPAs)}, which model sequences via compatibility in a shared semantic space.

In vision, models like \textbf{I-JEPA}~\citep{assran2023selfsupervisedlearningimagesjointembedding} and \textbf{V-JEPA}~\citep{bardes2024revisitingfeaturepredictionlearning} learn by reconstructing masked representations rather than raw inputs. This shifts supervision from low-level pixel-perfect reconstruction to high-level semantic consistency.

\textbf{Data2vec}~\citep{baevski2022data2vecgeneralframeworkselfsupervised} uses a similar approach across 3 separate domains of NLP, vision and audio by predicting contextual embeddings of masked inputs using a self-distillation loss. Without predicting tokens explicitly, it achieves competitive results across all 3 domains of language, vision, and speech.

\textbf{LLM-JEPA}~\citep{huang2025llmjepalargelanguagemodels} extends JEPA principles directly to language model training by treating different modalities (text descriptions and code) as views of the same underlying knowledge. Unlike traditional NTP objectives, LLM-JEPA learns representations by ensuring that semantically equivalent text-code pairs can predict each other in embedding space, while preventing dimensional collapse. The primary limitation of LLM-JEPA is its reliance on naturally paired multi-view data (such as documentation-code pairs), which restricts its application compared to standard self-supervised objectives. Despite this constraint, the method achieves significant improvements over baseline finetuning across multiple benchmarks (NL-RX, GSM8K, Spider) and model families (Llama3, Gemma2), demonstrating that JEPA's success in vision can transfer to language when suitable multi-view datasets exist.

\subsection{Failure Modes}
But these benefits come with trade-offs:

\begin{itemize}[leftmargin=1.5em, noitemsep]
    \item \textbf{No explicit generation:} Difficult to adapt to generative tasks without auxiliary decoders.
    \item \textbf{Representation drift:} Risk of collapse or over-smoothing during distillation.
    \item \textbf{Evaluation mismatch:} Embedding losses do not correlate cleanly with generative quality.
\end{itemize}

\subsection{Discussion}

Together, SSMs and JEPAs demonstrate two orthogonal routes away from NTP: one architectural, the other representational. While promising, both remain early-stage and face barriers in interpretability, generalization, and integration into practical generation systems.

\section{Conclusion \& Future Directions}

Next-Token Prediction (NTP) has long been the foundational paradigm for large-scale language modeling based on Transformers. While it enables impressive fluency, a growing body of literature highlights that NTP is also a bottleneck, with its myopic, token-by-token nature contributing to failures in long-term planning, computational efficiency, and semantic abstraction. In response, a rich ecosystem of alternatives has emerged. This survey has categorized these approaches into five main families: Multi-Token Prediction (MTP), Plan-then-Generate (PtG), Latent Reasoning (LR), Continuous Generation Approaches (CG), and Non-Transformer Architectures (NTA).

\subsection{Future Research Directions}
\label{sec:future-work}

\paragraph{Multi-Token Prediction (MTP).}
Based on this survey, we see two opportunities in this context. First, \emph{predict farther ahead}: study how performance changes as we increase $k$ (the number of future tokens), using simple schedules (start small, grow $k$) and checks for when extra lookahead stops helping. 

Secondly continue to \emph{use MTP as an auxillary training method}: keep future-token heads during training to teach lookahead, even if only next-token decoding is used at test time. 

However, it is important to note that a \emph{fixed} $k$ can never deliver truly global reasoning; longer documents and plans will always exceed any fixed horizon. Promising ideas include variable-length or adaptive horizons, and hierarchical heads that look ahead at different scales (phrases, sentences, sections).

\paragraph{Plan-then-Generate (PtG).}
\emph{Diverse planners}: move beyond generic embeddings to planners that capture discourse structure, key entities, or task steps, with evaluation focused on whether the generated text actually follows the plan. Secondly, further \emph{hybrid architectures} of different planning modules should be explored, along with systematic studies of \emph{adaptable time horizons}—plans that can flex between sentence-level, paragraph-level, or document-level guidance depending on task requirements.

\paragraph{Latent Reasoning.}
\emph{Diverse Architectures} that separate a small “reasoning core” from a lighter decoder, so we can add more internal thinking steps when needed. Secondly, improved \emph{Encoders/decoders} that faithfully move between text and latent space, with simple checks that meaning is preserved. Thirdly, \emph{non-Transformer models}, including state-space models and JEPA-style training, to predict in representation space rather than token space.

\paragraph{Continuous Generation Approaches.}
\emph{Compositional strategies}: combine diffusion with higher-level planning, for example by generating an outline of sections, entities, or steps and then refining the wording through a small number of denoising steps. This ensures the global structure is respected while keeping generation efficient. 

Secondly, \emph{memory-augmented diffusion}: pair diffusion with state-space memory to store long-context summaries in a recurrent module, while keeping attention over short windows. Diffusion can then refine current spans conditioned on this memory. Open questions include adaptive step counts, block-wise refinement with cache reuse, and training setups that balance global constraints (factuality, style) with local fluency.

\section*{Limitations}

This survey aims to provide a broad and structured overview of emerging alternatives to NTP in transformer models, despite some limitations. First, given the rapid pace of research in this area, our coverage may not include the most recent or unpublished advances, particularly those in proprietary or closed-source settings. Second, although we categorize alternatives along high-level axes (e.g., planning, token granularity, decoding strategy), the boundaries between categories are sometimes blurred. For example, some diffusion-based models also implement PtG, and some multi-token predictors incorporate latent semantic reasoning.

Moreover, our analysis focuses primarily on high-level paradigms that challenge the generative process of LLMs. Less emphasis is given to tokenization approaches, output head optimizations, or architectural innovations in the intermediate layers, which could meaningfully challenge the NTP paradigm. Finally, we do not present empirical benchmarking of these alternatives under controlled conditions; instead, we rely on results reported by the original authors.

\section*{Acknowledgments}
Portions of this manuscript were revised with the assistance of an AI-based editing tool (e.g., ChatGPT) for grammar and clarity. The authors are solely responsible for the content.

The visualizations used in this paper were created using Miro and Lucidchart.

\bibliography{main}

\end{document}